# Extrapolability Improvement of Machine Learning-Based Evapotranspiration Models via Domain-Adversarial Neural Networks

Haiyang Shi[1,2]

[1] Department of Civil and Environmental Engineering, University of Illinois at Urbana-Champaign, Urbana, IL 61801, USA

[2] State Key Laboratory of Desert and Oasis Ecology, Xinjiang Institute of Ecology and Geography, Chinese Academy of Sciences, Urumqi, 830011, China

Correspondence: Haiyang Shi (haiyang@illinois.edu)

## Abstract

Machine learning-based hydrological prediction models, despite their high accuracy, face limitations in extrapolation capabilities when applied globally due to uneven data distribution. This study integrates Domain-Adversarial Neural Networks (DANN) to improve the geographical adaptability of evapotranspiration (ET) models. By employing DANN, we aim to mitigate distributional discrepancies between different sites, significantly enhancing the model's extrapolation capabilities. Our results show that DANN improves ET prediction accuracy with an average increase in the Kling-Gupta Efficiency (KGE) of 0.2 to 0.3 compared to the traditional Leave-One-Out (LOO) method. DANN is particularly effective for isolated sites and transition zones between biomes, reducing data distribution discrepancies and avoiding low-accuracy predictions. By leveraging information from data-rich areas, DANN enhances the reliability of global-scale ET products, especially in ungauged regions. This study highlights the potential of domain adaptation techniques to improve the extrapolation and generalization capabilities of machine learning models in hydrological studies.

# 1 Introduction

With the accumulation of hydrological observations and the development of various big data technologies, machine learning-based hydrological prediction models have become increasingly popular, demonstrating high spatiotemporal prediction capabilities and accuracy (Reichstein et al., 2019). In such as streamflow and evapotranspiration (ET) simulations, machine learning has shown good performance (Feng et al., 2020; Jung et al., 2019; Nearing et al., 2024; Schnier & Cai, 2014; Shi et al., 2022a; X. Zeng et al., 2021). However, the current validation and evaluation of machine learning-based models are mostly based on gauged stations and catchments, which are unevenly distributed globally (Beck et al., 2016). In economically underdeveloped and remote areas, available observational data remain sparse (Hrachowitz et al., 2013) and it has raised concerns about the global extrapolability of process-based and data-driven hydrological models. How to transfer models built on data from well-gauged regions to ungauged regions remains a challenge.

For process-based hydrological models, extrapolability is often assessed by using hydrological similarity (Hrachowitz et al., 2013; Wagener et al., 2007) between ungauged catchments and gauged catchments as a basis for model parameter transfer (Beck et al., 2016). This approach assumes a mapping relationship from catchment characteristics to model parameters and subsequently to simulation accuracy. For example, previous studies found that using parameters transplanted from the 10 most similar 'donor' catchments provided more accurate predictions than using parameters calibrated from global catchments (Beck et al., 2016). In machine learning-based approaches, there have been relatively few explicit studies on such optimal input data selection strategies similarly. This may be due to the widespread belief that machine learning can automatically select similar data for simulation and prediction. Leave-one-site-out (LOO) cross-validation (Jung et al., 2019; Shi et al., 2022b, 2022a; Tramontana et al., 2016; J. Zeng et al., 2020) is a commonly used method to evaluate extrapolability for such models using all data for training and has been widely applied in cross-region ET simulations (Jung et al., 2019; Shi et al., 2022a). By leaving out the data of one site at a time during training as the test set, it comprehensively evaluates the model's prediction performance at ungauged sites.

However, data-driven models built using all available data are not always superior to those built using data subsets based on specific plant functional types (PFTs) (Shi et al., 2023). For some test sets from sites with specific PFTs, constructing the training set using data from corresponding PFT sites can achieve higher accuracy, as it avoids confusion from data of other types (Shi et al., 2023; Zhang et al., 2021). This approach is analogous to the use of the most similar 'donor' catchments for model parameter transfer in process-based hydrological models (Beck et al., 2016). Selecting input data from sites that are similar in hydrological and ecological characteristics may benefit the prediction accuracy for specific target sites (Shi et al., 2023). However, it is also recognized that this approach has not yet achieved optimal results. Whether the similarity obtained from empirical knowledge or hydrological and ecological characteristics matches well with model transferability remains uncertain. In machine learning-based runoff and flood prediction, while catchment properties are correlated with model performance, an accurate mapping relationship between model accuracy and sets of catchment attributes has yet to be established (Nearing et al., 2024).

Domain Adaptation (DA) is a machine learning technique aimed at adapting a model from a source domain to a target domain, even when there are significant differences in the data distributions between the two domains (Ben-David et al., 2006; Farahani et al., 2021; Zhuang et al., 2021). Model performance can be enhanced in new environments, improving generalization capability and prediction accuracy without requiring a large amount of labelled data from the target domain, especially in unsupervised DA backpropagation training (Ganin & Lempitsky, 2015). Various methods exist for DA, including instance re-weighting, feature mapping, and adversarial training (Ganin et al., 2016). Adversarial training involves training the model in an adversarial manner to make it more robust when facing different tasks (Bai et al., 2021). The goal is to make the features generated by the feature extractor indistinguishable between the source and target domains, thereby achieving domain adaptation. For example, the Domain-Adversarial Neural Network (DANN) (Ganin et al., 2016; Zhuang et al., 2021) can reduce data distribution discrepancies between domains through adversarial interaction between the domain classifier and the feature extractor. We recognize that ET simulation is a typical scenario for applying DA (Shi, 2024). Due to variations in meteorological conditions, vegetation types, and soil characteristics across different geographical locations, traditional machine learning models often suffer from performance degradation when making cross-site predictions. DA has the potential to improve performance in this condition (Shi, 2024) by leveraging the rich data and information from the source domain (e.g., certain flux sites) and adjusting the model to perform better in the target domain (e.g., new flux sites).

Therefore, by using DANN, we evaluated the role of DA in improving the extrapolability of ET simulation models. To investigate whether DA-based models truly have an advantage, we used the Random Forests model (RF) combining the LOO method (frequently used and showed relatively high model performance (Shi et al., 2022a)) for performance comparison. Flux station data from FLUXNET2015 along with various other predictor variables were used as the training set. This study aims to provide useful guidance on how to enhance the cross-site and cross-regional extrapolability of machine learning models in hydrological studies, including ET prediction.

## 2 Methodology

### 2.1 Flux data and predictors

ET observational data are derived from latent heat observations in the FLUXNET2015 flux dataset (Pastorello et al., 2020) and its meteorological observations are used as inputs. Other satellite remote sensing data include estimates of downward shortwave radiation (RSDN) and leaf area index (LAI) and they were extracted at a 500x500 m scale surrounding the site location by Google Earth Engine. To measure the static or multi-year average ecological and hydrological properties of the sites, we extracted data from a site-scale dataset (Migliavacca et al., 2021) for FLUXNET2015 sites, including GSmax, G1, GPPsat, Nmass, LAImax, MAT, MAP, CSWI, and Hc (Table 1). Soil sand content is derived from SoilGrid data. A total of 129 site datasets were selected (Fig. 1), with the forest sites being the most numerous (71 sites). To reduce computational cost, we used data from the year with the most complete records for each site for training and validation. This approach enhances the evaluation's cross-site fairness when comparing the extrapolatability, making our evaluation more focused on geographic variability.

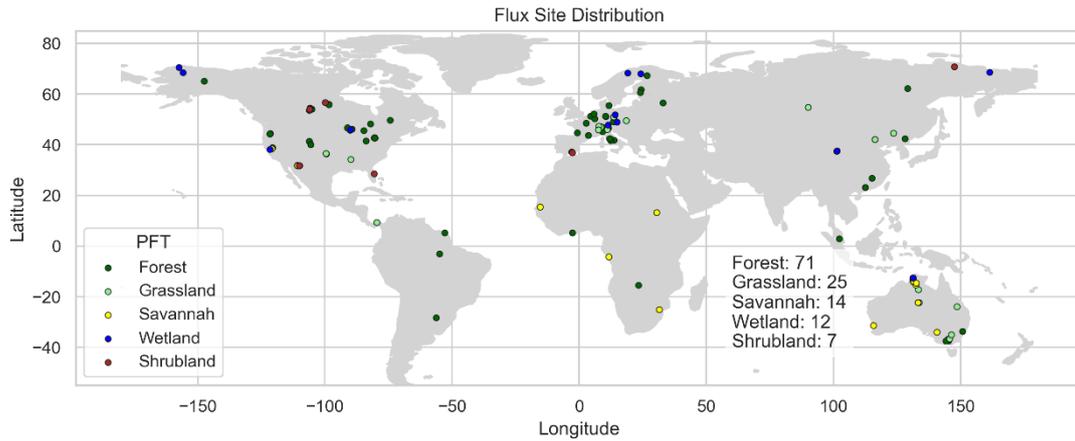

**Figure 1.** Flux sites used for model training.

**Table 1** Predictors used in the model training.

| Predictor | Description | Dynamic/static | Source |
|---|---|---|---|
| TA | Daily average air temperature | Dynamic | FLUXNET2015 (Pastorello et al., 2020) |
| TA_range | Daily air temperature range | | |
| VPD | Vapor pressure deficit (calculated from daily average temperature and dew point temperature) | | |
| WS | Wind speed | | |
| RSDN | Downward shortwave radiation | | BESS (Ryu et al., 2018) |
| LAI | Leaf area index | | MODIS imagery: MCD15A3H |
| SSM | Surface soil moisture estimates | | Ref. (Han et al., 2023) |
| GSmax | Maximum surface conductance | Static or multi-year mean value | Ref. (Migliavacca et al., 2021) |
| G1 | Stomatal slope | | |
| GPPsat | Gross primary productivity at light saturation | | |
| MAT | Mean annual temperature | | |
| MAP | Mean annual precipitation | | |
| Hc | Canopy height | | |

| sand_frac | Soil sand content of 5-15 cm depth | | SoilGrid250m (Hengl et al., 2017) |

## 2.2 LOO-based RF modeling

RF is a commonly used machine learning method, and it has been utilized as a predictive algorithm in various ET predictions and products (Jung et al., 2019; Shi et al., 2022a) and has demonstrated feasibility and relatively high accuracy (Shi et al., 2022a). Here, RF-based accuracy is used for comparison with DANN. The RF modeling is conducted in two ways: using LOO and using only hydrologically similar sites as training data (as introduced in Section 2.3). The LOO method is a widely used validation technique to assess the generalization ability of models in flux tower-based ET simulations. In this approach, data from multiple flux towers, including latent heat flux, meteorological observations, and other relevant variables are collected and grouped by site and year. For each iteration of the LOO process, data from one site is set aside as the test set, while data from all other sites are used to train the model. The trained model then predicts ET for the test site, and the predictions are compared to the observed values. This process is repeated for each site, allowing for a comprehensive assessment of the model's performance across different sites. It provides a relatively fair basis for comparing different models to determine the best-performing model across diverse site conditions.

In this study, we set the number of trees in the RFs to 50 and set other RF parameters as the default values of the 'RandomForestRegressor' function of the 'scikit-learn' Python package. We used Kling-Gupta Efficiency (KGE) (Gupta et al., 2009; Knoben et al., 2019) as the model performance metric to measure the consistency between simulated and observed ET. KGE addresses several shortcomings in NSE and is increasingly used for model calibration and evaluation:

$$KGE = 1 - \sqrt{(r-1)^2 + \left(\frac{\sigma_{sim}}{\sigma_{obs}} - 1\right)^2 + \left(\frac{\mu_{sim}}{\mu_{obs}} - 1\right)^2} \qquad (1)$$

where $r$ represents the linear correlation between observations and simulations, $\sigma_{obs}$ represents the standard deviation of observation values, $\sigma_{sim}$ represents the standard deviation of simulation values, $\mu_{sim}$ represents the mean value of simulation values, and $\mu_{obs}$ represents the mean value of observation values. KGE = 1 indicates perfect consistency between simulations and observations, and higher KGE values correspond to higher consistency between simulation and observation.

**2.3 Hydrological similarity-based RF modeling**

The RF modeling based on hydrological similarity referenced the parameter transfer schemes from process-based hydrological models (Beck et al., 2016). For instance, in process-based models, parameter schemes calibrated from the 10 most similar 'donor' catchments based on hydrological properties are transferred to an ungauged basin (Beck et al., 2016). In this study, for each target site, the training set comprised the 10 most similar sites with the same PFT. This similarity was calculated simply using the Euclidean distance of normalized MAT, MAP, and Hc. MAT and MAP can effectively characterize the hydrological and climatic properties of a site, while Hc can represent the vegetation's hydraulic properties, such as rooting depth and drought tolerance (Giardina et al., 2018). For the Shrubland type, since the number of sites was fewer than 10, all available Shrubland sites were used as the training set. Due to the reduced number of sites and data volume in the training set, the number of trees in the RF model was set to 20, while other parameters remained unchanged. The model performance for each site was also evaluated using the KGE metric.

**2.4 DANN construction**

We employed a DANN (Ganin et al., 2016; Zhuang et al., 2021) to enhance the accuracy of cross-site ET prediction by mitigating distributional discrepancies between different sites. The DANN architecture (Fig. 2) comprises three main components: a feature extractor, an ET predictor, and a domain classifier. The feature extractor, consisting of two fully connected layers with ReLU activations, extracts relevant features from the input data. The ET predictor, also composed of fully connected layers with ReLU activations, performs the regression task of predicting ET values. The domain classifier, designed with a similar neural network structure, distinguishes whether the features are from the source domain or the target domain. This adversarial training approach encourages the feature extractor to learn domain-invariant features, which enhances the model's generalization to new sites.

The training process involves minimizing a total loss that combines regression and domain losses (Ganin et al., 2016; Zhuang et al., 2021). The regression loss $L_r$ is calculated using Mean Squared Error (MSE), which measures the difference between predicted and actual ET values. The domain loss $L_d$ uses Cross-Entropy Loss to evaluate the domain classifier's performance in distinguishing between source and target domains:

$$L_d = -\frac{1}{N}\sum_{i=1}^{N}[d_i \log(\hat{d_i}) + (1 - d_i)\log(1 - \hat{d_i})] \qquad (2)$$

where $d_i$ is the domain label (0 for source, 1 for target) and $\hat{d_i}$ is the predicted domain label. N is the number of samples. The total loss $L_{total}$ is the sum of the regression loss and the domain loss, weighted by a dynamically adjusted parameter $\lambda$:

$$L_{total} = L_r + \lambda \cdot L_d \qquad (3)$$

The weight parameter $\lambda$ is adjusted during training based on the training progress to balance the two losses:

$$\lambda = \frac{2}{1+\exp(-10 \cdot progress)} - 1 \qquad (4)$$

where *progress* represents the normalized training progress from 0 to 1. The model is trained using the Adam optimizer with a learning rate of 0.001, and the training loop iterates over 50 epochs. Each epoch dynamically adjusts $\lambda$ to balance the importance of domain and regression losses. After training, the model's performance is also evaluated on the test set using the KGE metric.

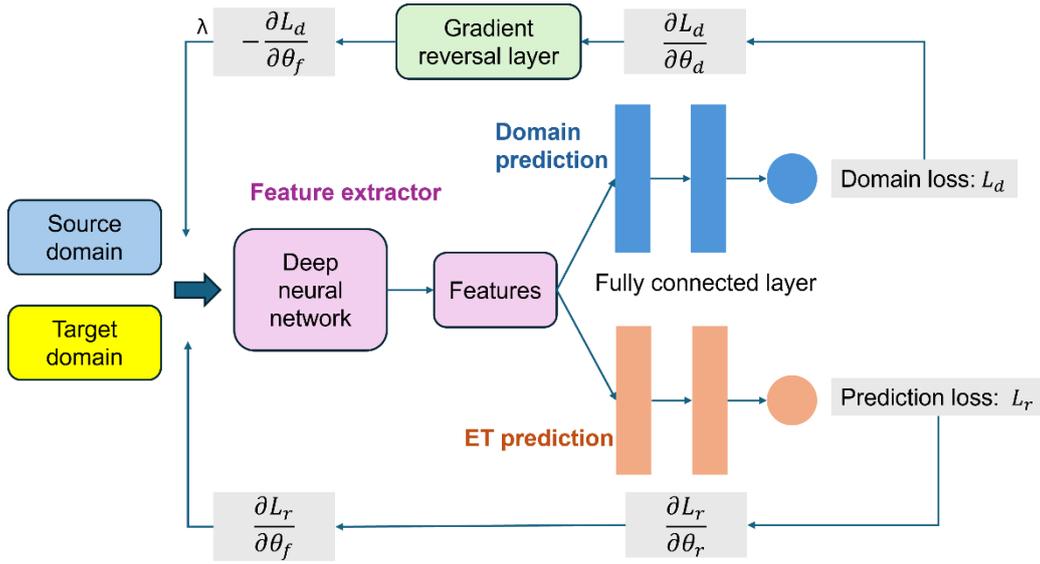

**Figure 2.** The architecture of the DANN for ET prediction. $\theta_f$, $\theta_d$, and $\theta_r$ denote trainable model weights in the feature extractor, domain classifier (for domain prediction), and ET predictor (for ET prediction), respectively. $L_r$ denotes the regression loss and $L_d$ denotes the domain loss.

## 3 Results

The KGE values for DANN exceeded 0.8 at most sites, whereas many sites exhibited KGE values below 0.5 with LOO and PFT-specific modeling approaches (Fig. 3). In the PFT-specific modeling approach, some sites even recorded negative KGE values, indicating model failure and an inability to explain data variability. Compared to LOO, the average increase in KGE for DANN ranged from 0.2 to 0.3, representing a significant improvement in accuracy. Compared to LOO, DANN demonstrated a median KGE improvement of over 0.2 for forests, grasslands, and savannahs (Fig. 4). For wetlands and shrublands, the median KGE improvement was slightly lower but still exceeded 0.1. Notably, except for the shrubland type, the lower limit of the KGE values for DANN exceeded 0.7. In pairwise comparisons, DANN showed a slightly lower KGE than LOO at only three sites, while significantly improving accuracy at other sites. For shrublands, DANN's 25th percentile KGE was lower than 0.8. This indicates that DANN significantly improves accuracy by ensuring that most sites achieve a KGE over 0.8, effectively avoiding low KGE occurrences. In contrast, LOO generally exhibited higher KGE values than the PFT-specific modeling approach across most sites. However, the PFT-specific approach occasionally outperformed LOO at certain sites, such as some forest sites, where the KGE was higher by 0.1 to 0.3.

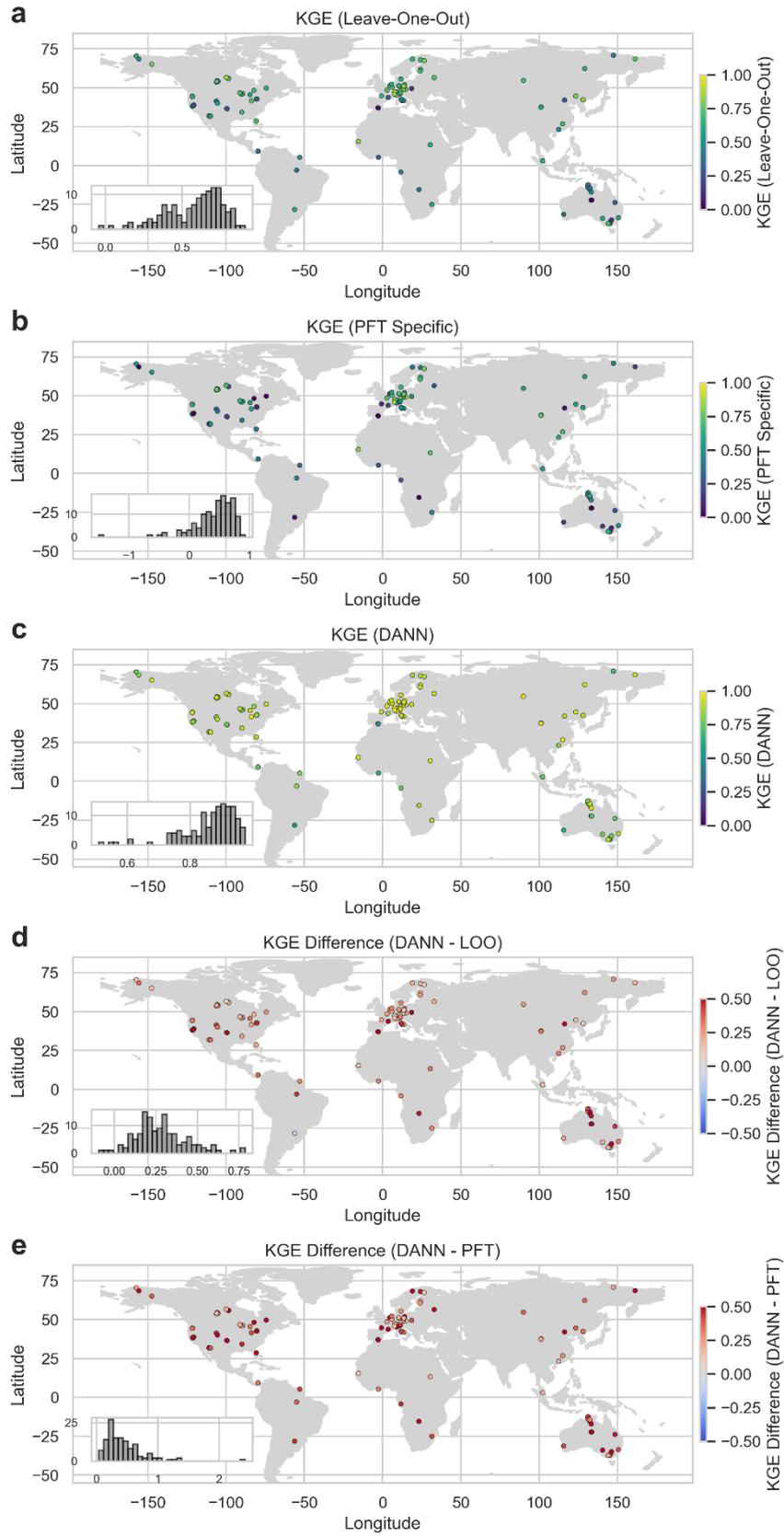

**Figure 3.** The distribution of KGE of the flux sites (a-c) under the three modeling approaches (i.e., LOO, 'PFT specific', and DANN), as well as the accuracy advantage of DANN compared to the other two methods (d-e). 'PFT specific' refers to the modeling approach based on the 10 most similar sites within the same PFT.

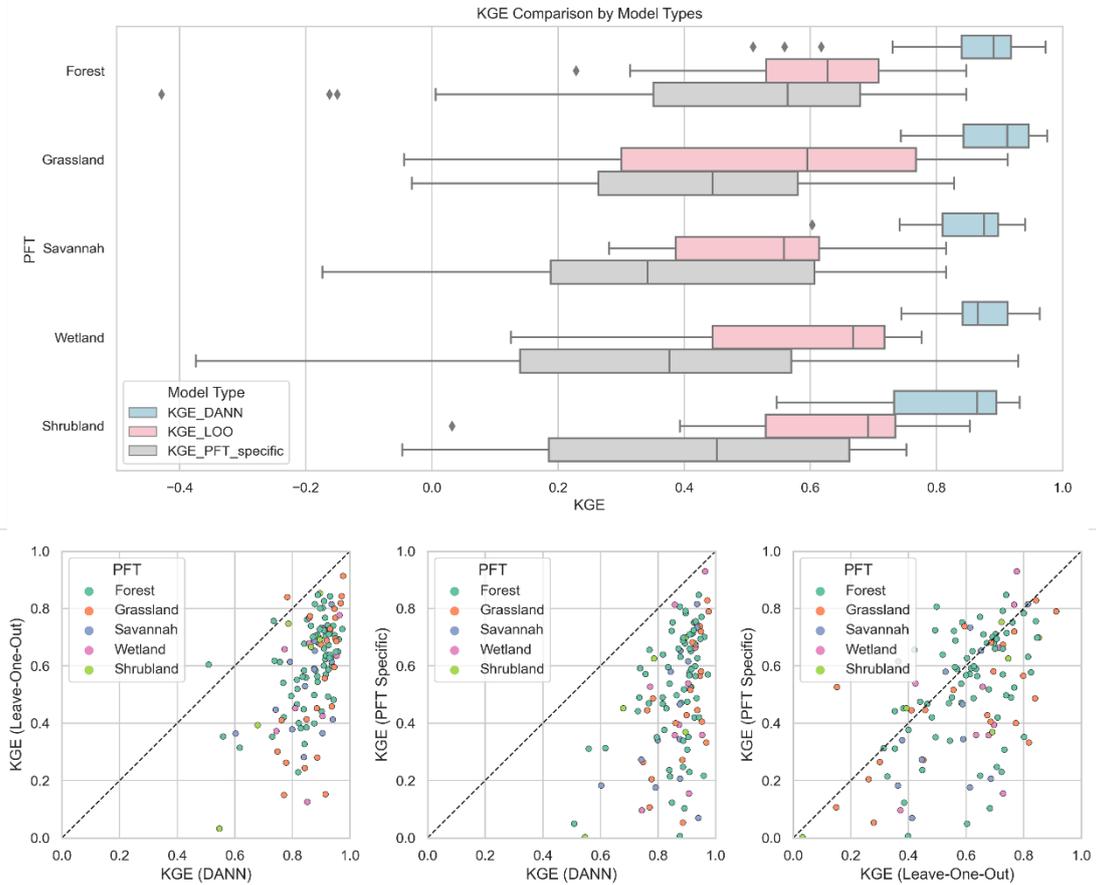

**Figure 4.** The KGE differences of the three modeling approaches across various PFT sites (upper part) and their pairwise comparisons (lower part).

The distribution of biomes corresponding to sites where the DANN model shows improvement in KGE compared to LOO also exhibits a pattern (Fig. 5). Sites with smaller KGE improvements are mostly found in biomes such as Woodland Shrubland, Temperate Forest, and Boreal Forest, where flux tower distributions are more concentrated. In contrast, greater KGE improvements are observed at more isolated sites (with sparse surrounding site distribution). This indicates that DANN can enhance model performance by learning domain-invariant features, which is particularly useful in scenarios with limited or sparse training data. Isolated sites benefit from DANN's transfer learning capabilities. By adapting features learned from data-rich regions, DANN can make more informed predictions at isolated sites, rather than relying solely on local data.

Additionally, sites with significant KGE improvements are frequently located in transition zones among Subtropical Desert, Temperate Grassland Desert, Woodland Shrubland, and Tropical Forest Savannah, as well as between Subtropical Desert and Tropical Forest Savannah. In these transition zones, DANN's DA mechanism proves effective. By reducing data distribution discrepancies between different PFTs, DANN can enhance model performance at sites in these transition zones, whereas traditional models may struggle with the mixed characteristics of PFTs. Transition zones can benefit from DANN's generalization ability. Given the mixed features of these regions, a model capable of generalizing well across different domains can provide more accurate predictions. Overall, DANN's DA capability makes it particularly suitable for improving ET prediction accuracy in regions with sparse site distribution or heterogeneous conditions in the predictor space.

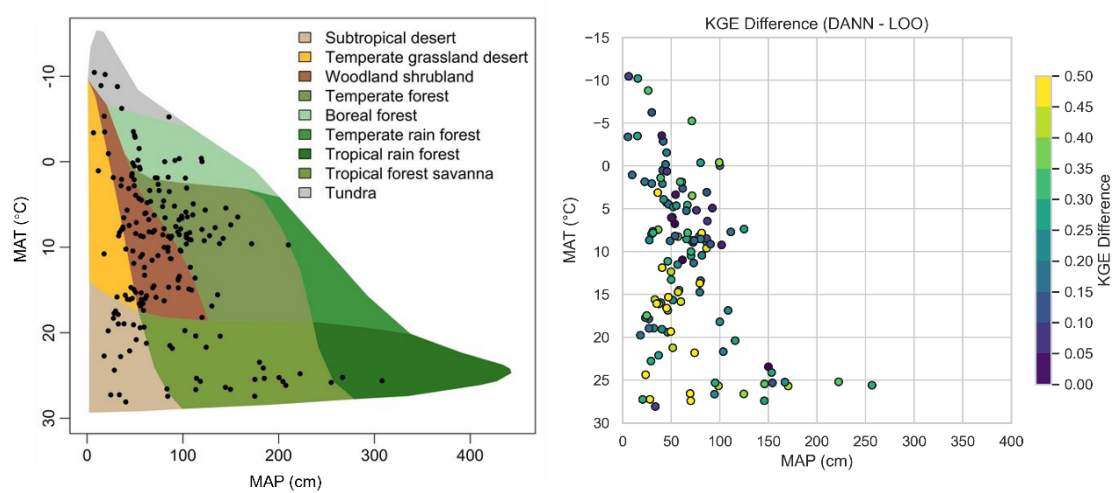

**Figure 5.** The distribution of the KGE improvement of DANN compared to LOO (right part) and the distribution of the biomes of the corresponding flux sites (left part) across MAT and MAP (modified from the ref. (Migliavacca et al., 2021)).

## 4 Discussions

In previous global-scale data-driven hydrological prediction models, despite the relatively high accuracy achieved by models such as RF (Jung et al., 2019; Shi et al., 2022a), the extrapolation capability remains limited. This study applies DA techniques commonly used in computer vision science and image classification science to improve the model's adaptation to geographical variability. Specifically, we employed DANN and significantly enhanced the extrapolation ability of the ET prediction model.

Compared to the traditional LOO method, DANN can increase the KGE of ET predictions by an average of 0.2 to 0.3, while largely avoiding instances of low accuracy. When applied on a global scale, DANN is expected to greatly improve the reliability of current global-scale ET products (Jung et al., 2019; Martens et al., 2017). This improvement is particularly beneficial for isolated locations in the predictor space. DANN achieves higher accuracy for these locations through data augmentation by leveraging information from data-rich areas. This capability is potentially very useful for global-scale predictions, especially in ungauged regions lacking observational data for calibration and validation. For sites located in transition zones between different biomes, DANN can reduce data distribution discrepancies (Farahani et al., 2021) between different PFTs or biomes, thus avoiding the confusion that traditional models might encounter. In contrast to LOO, the method of selecting sites' data based on hydrological similarity did not show a comprehensive improvement in accuracy. Although it can achieve higher accuracy at some sites by reducing heterogeneity within the training set, it often results in low KGE values. It may be because selecting data from only the 10 most similar sites for training would lead to a reduction in the amount of data, and would also exclude useful information that could be provided by other less similar sites (Shi et al., 2023). It suggests that the selection of the few most similar catchments in the parameter transfer scheme for process-based hydrological models (Beck et al., 2016; Guo et al., 2021) is not necessarily applicable to machine learning-based models.

Overall, as a DA technique, DANN can comprehensively improve the extrapolation capability of current ET prediction models, thereby potentially enhancing the accuracy of global-scale ET products significantly. Other machine learning models in hydrology and ecology, which involve geographical heterogeneity, should also consider using DA techniques to enhance their extrapolation and generalization capabilities in large-scale applications (Shi, 2024).

# 5 Conclusion

In ET prediction, compared to traditional LOO methods that use all available data directly for training and methods that select similar sites' data based on hydrological similarity, DANN can significantly improve extrapolation capability and model performance (with an average KGE improvement of 0.2 to 0.3). DANN is particularly useful for isolated sites and sites located in transition zones between different biomes. Future global-scale ET products based on DANN are expected to achieve higher accuracy and reliability in ungauged regions.

# References

Bai, T., Luo, J., Zhao, J., Wen, B., & Wang, Q. (2021, April 20). Recent Advances in Adversarial Training for Adversarial Robustness. arXiv. https://doi.org/10.48550/arXiv.2102.01356

Beck, H. E., van Dijk, A. I. J. M., de Roo, A., Miralles, D. G., McVicar, T. R., Schellekens, J., & Bruijnzeel, L. A. (2016). Global-scale regionalization of hydrologic model parameters. *Water Resources Research*, *52*(5), 3599–3622. https://doi.org/10.1002/2015WR018247

Ben-David, S., Blitzer, J., Crammer, K., & Pereira, F. (2006). Analysis of representations for domain adaptation. *Advances in Neural Information Processing Systems*, *19*.

Farahani, A., Voghoei, S., Rasheed, K., & Arabnia, H. R. (2021). A Brief Review of Domain Adaptation. In R. Stahlbock, G. M. Weiss, M. Abou-Nasr, C.-Y. Yang, H. R. Arabnia, & L. Deligiannidis (Eds.), *Advances in Data Science and Information Engineering* (pp. 877–894). Cham: Springer International Publishing. https://doi.org/10.1007/978-3-030-71704-9_65

Feng, D., Fang, K., & Shen, C. (2020). Enhancing Streamflow Forecast and Extracting Insights Using Long-Short Term Memory Networks With Data Integration at Continental Scales. *Water Resources Research*, *56*(9), e2019WR026793. https://doi.org/10.1029/2019WR026793

Ganin, Y., & Lempitsky, V. (2015). Unsupervised domain adaptation by backpropagation (pp. 1180–1189). Presented at the International conference on machine learning, PMLR.

Ganin, Y., Ustinova, E., Ajakan, H., Germain, P., Larochelle, H., Laviolette, F., et al. (2016). Domain-Adversarial Training of Neural Networks. *Journal of Machine Learning Research*, *17*(59), 1–35.

Giardina, F., Konings, A. G., Kennedy, D., Alemohammad, S. H., Oliveira, R. S., Uriarte, M., & Gentine, P. (2018). Tall Amazonian forests are less sensitive to precipitation variability. *Nature Geoscience*, *11*(6), 405–409. https://doi.org/10.1038/s41561-018-0133-5

Guo, Y., Zhang, Y., Zhang, L., & Wang, Z. (2021). Regionalization of hydrological modeling for predicting streamflow in ungauged catchments: A comprehensive review. *WIREs Water*, *8*(1), e1487. https://doi.org/10.1002/wat2.1487

Gupta, H. V., Kling, H., Yilmaz, K. K., & Martinez, G. F. (2009). Decomposition of the mean squared error and NSE performance criteria: Implications for improving hydrological modelling. *Journal of Hydrology*, *377*(1), 80–91. https://doi.org/10.1016/j.jhydrol.2009.08.003

Han, Q., Zeng, Y., Zhang, L., Wang, C., Prikaziuk, E., Niu, Z., & Su, B. (2023). Global long term daily 1 km surface soil moisture dataset with physics informed machine learning. *Scientific Data*, *10*(1), 101. https://doi.org/10.1038/s41597-023-02011-7

Hengl, T., Mendes de Jesus, J., Heuvelink, G. B., Ruiperez Gonzalez, M., Kilibarda, M., Blagotić, A., et al. (2017). SoilGrids250m: Global gridded soil information based on machine learning. *PLoS One*, *12*(2), e0169748.

Hrachowitz, M., Savenije, H. H. G., Blöschl, G., McDonnell, J. J., Sivapalan, M., Pomeroy, J. W., et al. (2013). A decade of Predictions in Ungauged Basins (PUB)—a review. *Hydrological Sciences Journal*, *58*(6), 1198–1255. https://doi.org/10.1080/02626667.2013.803183

Jung, M., Koirala, S., Weber, U., Ichii, K., Gans, F., Camps-Valls, G., et al. (2019). The FLUXCOM ensemble of global land-atmosphere energy fluxes. *Scientific Data*, *6*(1), 74. https://doi.org/10.1038/s41597-019-0076-8

Knoben, W. J. M., Freer, J. E., & Woods, R. A. (2019). Technical note: Inherent benchmark or not? Comparing Nash–Sutcliffe and Kling–Gupta efficiency scores. *Hydrology and Earth System Sciences*, *23*(10), 4323–4331. https://doi.org/10.5194/hess-23-4323-2019

Martens, B., Miralles, D. G., Lievens, H., Van Der Schalie, R., De Jeu, R. A., Fernández-Prieto, D., et al. (2017). GLEAM v3: Satellite-based land evaporation and root-zone soil moisture. *Geoscientific Model Development*, *10*(5), 1903–1925.

Migliavacca, M., Musavi, T., Mahecha, M. D., Nelson, J. A., Knauer, J., Baldocchi, D. D., et al. (2021). The three major axes of terrestrial ecosystem function. *Nature*, *598*(7881), 468–472. https://doi.org/10.1038/s41586-021-03939-9

Nearing, G., Cohen, D., Dube, V., Gauch, M., Gilon, O., Harrigan, S., et al. (2024). Global prediction of extreme floods in ungauged watersheds. *Nature*, *627*(8004), 559–563. https://doi.org/10.1038/s41586-024-07145-1

Pastorello, G., Trotta, C., Canfora, E., Chu, H., Christianson, D., Cheah, Y.-W., et al. (2020). The FLUXNET2015 dataset and the ONEFlux processing pipeline for eddy covariance data. *Scientific Data*, *7*(1), 225. https://doi.org/10.1038/s41597-020-0534-3

Reichstein, M., Camps-Valls, G., Stevens, B., Jung, M., Denzler, J., Carvalhais, N., & Prabhat. (2019). Deep learning and process understanding for data-driven Earth system science. *Nature*, *566*(7743), 195–204. https://doi.org/10.1038/s41586-019-0912-1

Ryu, Y., Jiang, C., Kobayashi, H., & Detto, M. (2018). MODIS-derived global land products of shortwave radiation and diffuse and total photosynthetically active radiation at 5 km resolution from 2000. *Remote Sensing of Environment*, *204*, 812–825. https://doi.org/10.1016/j.rse.2017.09.021

Schnier, S., & Cai, X. (2014). Prediction of regional streamflow frequency using model tree ensembles. *Journal of Hydrology*, *517*, 298–309. https://doi.org/10.1016/j.jhydrol.2014.05.029

Shi, H. (2024, March 17). Potential of Domain Adaptation in Machine Learning in Ecology and Hydrology to Improve Model Extrapolability. arXiv. https://doi.org/10.48550/arXiv.2403.11331

Shi, H., Luo, G., Hellwich, O., Xie, M., Zhang, C., Zhang, Y., et al. (2022a). Evaluation of water flux predictive models developed using eddy-covariance observations and machine learning: a meta-analysis. *Hydrology and Earth System Sciences*, *26*(18), 4603–4618. https://doi.org/10.5194/hess-26-4603-2022

Shi, H., Luo, G., Hellwich, O., Xie, M., Zhang, C., Zhang, Y., et al. (2022b). Variability and uncertainty in flux-site-scale net ecosystem exchange simulations based on machine learning and remote sensing: a systematic evaluation. *Biogeosciences*, *19*(16), 3739–3756. https://doi.org/10.5194/bg-19-3739-2022

Shi, H., Luo, G., Hellwich, O., He, X., Xie, M., Zhang, W., et al. (2023). Comparing the use of all data or specific subsets for training machine learning models in hydrology: A case study of evapotranspiration prediction. *Journal of Hydrology*, *627*, 130399. https://doi.org/10.1016/j.jhydrol.2023.130399

Tramontana, G., Jung, M., Schwalm, C. R., Ichii, K., Camps-Valls, G., Ráduly, B., et al. (2016). Predicting carbon dioxide and energy fluxes across global FLUXNET sites with regression algorithms. *Biogeosciences*, *13*(14), 4291–4313. https://doi.org/10.5194/bg-13-4291-2016

Wagener, T., Sivapalan, M., Troch, P., & Woods, R. (2007). Catchment Classification and

Hydrologic Similarity. *Geography Compass*, *1*(4), 901–931. https://doi.org/10.1111/j.1749-8198.2007.00039.x

Zeng, J., Matsunaga, T., Tan, Z.-H., Saigusa, N., Shirai, T., Tang, Y., et al. (2020). Global terrestrial carbon fluxes of 1999–2019 estimated by upscaling eddy covariance data with a random forest. *Scientific Data*, *7*(1). https://doi.org/10.1038/s41597-020-00653-5

Zeng, X., Schnier, S., & Cai, X. (2021). A data-driven analysis of frequent patterns and variable importance for streamflow trend attribution. *Advances in Water Resources*, *147*, 103799. https://doi.org/10.1016/j.advwatres.2020.103799

Zhang, C., Luo, G., Hellwich, O., Chen, C., Zhang, W., Xie, M., et al. (2021). A framework for estimating actual evapotranspiration at weather stations without flux observations by combining data from MODIS and flux towers through a machine learning approach. *Journal of Hydrology*, *603*, 127047. https://doi.org/10.1016/j.jhydrol.2021.127047

Zhuang, F., Qi, Z., Duan, K., Xi, D., Zhu, Y., Zhu, H., et al. (2021). A Comprehensive Survey on Transfer Learning. *Proceedings of the IEEE*, *109*(1), 43–76. https://doi.org/10.1109/JPROC.2020.3004555